\documentclass[10pt,twocolumn,letterpaper]{article}

\usepackage{cvpr}
\usepackage{times}
\usepackage{epsfig}
\usepackage{graphicx}
\usepackage{amsmath}
\usepackage{amssymb}
\usepackage{glossaries}
\usepackage{tabularx}
\usepackage{array}
\usepackage{subfig}
\newcolumntype{P}[1]{>{\centering\arraybackslash}p{#1}}

\usepackage[pagebackref=true,breaklinks=true,letterpaper=true,colorlinks,bookmarks=false]{hyperref}

\cvprfinalcopy 

\def\cvprPaperID{****} 

\begin{document}

\title{Slash or burn: Power line and vegetation classification for wildfire prevention}

\author{Austin Park, Farzaneh Rajabi, and Ross Weber\\
Stanford University\\
Department of Energy Resources Engineering
}

\maketitle

\begin{abstract}
   Electric utilities are struggling to manage increasing wildfire risk in a hotter and drier climate. Utility transmission and distribution lines regularly ignite destructive fires when they make contact with surrounding vegetation. Trimming vegetation to maintain the separation from utility assets is as critical to safety as it is difficult. Each utility has tens of thousands of linear miles to manage, poor knowledge of where those assets are located, and no way to prioritize trimming. Feature-enhanced convolutional neural networks (CNNs) have proven effective in this problem space. Histograms of oriented gradients (HOG) and Hough transforms are used to increase the salience of the linear structures like power lines and poles. Data is frequently taken from drone or satellite footage, but Google Street View offers an even more scalable and lower cost solution. This paper uses $1,320$ images scraped from Street View, transfer learning on popular CNNs, and feature engineering to place images in  one  of  three classes:  (1) no utility systems, (2) utility systems with no overgrown vegetation, or (3) utility systems with overgrown vegetation. The CNN output thus yields a prioritized vegetation management system and creates a geotagged map of utility assets as a byproduct. Test set accuracy with reached $80.15\%$ using \textit{VGG11} with a trained first layer and classifier, and a model ensemble correctly classified $88.88\%$ of images with risky vegetation overgrowth.
\end{abstract}

\section{Introduction}

In a hotter and more drought-prone world, wildfire risk will continue to increase. The situation is particularly dire for the Northern Californian investor-owned utility: "Cal Fire [determined] in May that PG\&E lines were the cause of several fires that killed at least 15 people and razed over 5,000 homes in the fall of 2017, including 12 instances in which it found the utility in violation of safety or maintenance procedures"~\cite{John-2019-Cal}. In 2018, the Camp Fire killed 85 people, destroyed the town of Paradise and deteriorated air pollution throughout Northern California for nearly a week. PG\&E is liable for up to \$17B from 2018's wildfires. Finding exact locations where vegetation and power lines intersect can help avoid these fires, which leads to significant savings in money, carbon emissions, and lives.

Electric utilities hoping to strategically reduce wildfire risk in their service territories immediately face three hurdles: First, they often have tens of thousands of linear transmission and distribution (T\&D) miles. PG\&E is responsible for roughly 106,000 miles of T\&D, much of which requires annual inspection ~\cite{John-2019-Cal}. Clearly, any vegetation management solution needs to scale.  Second, utilities lack comprehensive maps of their assets. They installed many of their electricity distribution systems over 60 years ago, and there are no records or employees left from that era. Third, there is no standardized and rigorous way to prioritize maintenance. "Utilities utilities typically rely on vegetation management contractors to decide how vegetation should be maintained" ~\cite{malashenko2018powergrid}. Inspection is currently done in two stages: A trained arborist identifies infringing vegetation, and then subcontracts a tree trimming service to actually remove the vegetation. The only form of prioritization comes from customers reporting outages and fires.

The scale of the problem makes a software solution highly appealing. Governor Newsom’s Strike Force released a report specifically asking for Artificial Intelligence-based visual recognition technology to analyze satellite imagery to determine fuel conditions and vegetation risks in proximity to utility lines," and "machine learning and automation inspections for increased safety assurance and regulatory compliance" ~\cite{newsom2018wildfires}. Public, private, and education sectors agree that machine learning solutions to vegetation management appear promising.

Google Street View images provides an even more accessible dataset than satellite imagery, and Street View images are available for nearly the entirety of PG\&E's service territory. Feature transformations such as the Histogram of Oriented Gradients (HOG) and Hough transformation have proven successful in identifying power lines and similar structure~\cite{li2008towards, fernandes2008real, Gubbi-2018-New}. Thus, in this work, a convolutional neural network (CNN) is trained using Google Street View images and these feature transformation as inputs, and the output is a classification decision determining whether or not the image contains utility assets and if they are considered at risk. An extremely useful byproduct of this approach is a geotagged list of utility assets. Last, by choosing classes with care, the CNN can output an effective a prioritization system.

This paper is structured as follows. Section~\ref{sec:2} describes recent computer vision approaches that have been applied to line detection. Section~\ref{sec:3} describes the methods used for training the CNN in the work. The dataset and feature transformations are described in Section~\ref{sec:4}, and Section~\ref{sec:5} shows the results of various networks trained for the classification task.

\section{Related Work} \label{sec:2}

Recently, computer vision schemes have been employed to classify images into different land cover classes such as building, grassland, dense vegetation, water body, barren land, road, and shadow~\cite{sameen2007class}. However, the task of wire detection remains an open challenge; wires are generally only a few pixels wide, can appear at any orientation and location, may appear non-linear due to low tension or poor picture stitching, and are hard to distinguish from other similar looking lines and edges.

Utility companies currently spend an enormous amount of time and money performing visual inspections on their transmission and distribution networks, using helicopters, unmanned aerial vehicles, and climbing robots. In addition to the expense and slow implementation, some of these solutions can be dangerous. Luckily, CNNs are making great strides in power line detection, using drone footage, Google Street View, and satellite images as data sources~\cite{Madaan-2018-Wire, liu2016insulator, Zhang-2018-Using, martinez2017PoLIS, siddiqui2018robust}. Detection of transmission lines relies on precisely locating objects of interest and their locations with respect to one another, making it an ideal problem for computer vision. Madann et al. and Siddiqui et al. focus on insulator detection using bounding boxes, with the latter attempting to detect defects ~\cite{Madaan-2018-Wire, siddiqui2018robust}. Zhang et al. show Google Street View can accurately perform object detection and geotag utility poles within a 10 meter radius ~\cite{Zhang-2018-Using}. 

To improve CNN performance in power line detection, many researchers are employing feature engineering \cite{Nguyen-2018-Automatic}. Previous approaches have included these features without CNNs, but they typically perform less successfully and robustly than CNNs. For example, Song et al. used a matched derivative variant called the first-order derivative of Gaussian. It works well when there are no other edges in the image, but cannot distinguish a street edge from a power line ~\cite{song2014power}. Two feature transforms that have been applied for power line detection are the Histogram of Oriented Gradients (HOG) and the Hough transform~\cite{li2008towards, fernandes2008real, Gubbi-2018-New}. The Hough transform extracts canny edges, performs a polar transform, and then finds line segments in the polar space. In the HOG transform, the distribution of directional gradients over local patches of the image are used collectively as a feature. Gubbi et. al employed this approach to detect power lines from satellite images~\cite{Gubbi-2018-New}.

\section{Methods} \label{sec:3}

\begin{figure*}[h!]
    \centering
    \includegraphics[width=\linewidth]{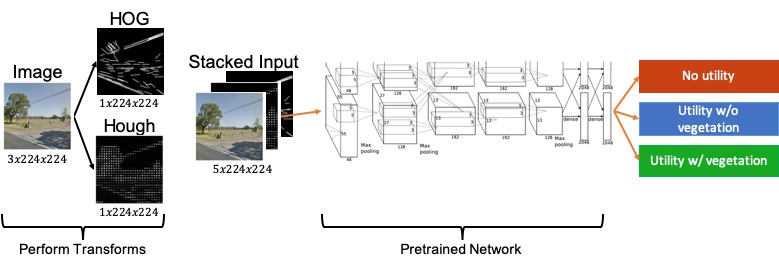}
    \caption{Chart detailing the process used for training a utility line classifier from Google Street View images. The HOG and Hough transforms are first applied and then stacked together with the original image as the input into a pretrained CNN.}
    \label{fig:net}
\end{figure*}

This work uses a feature-enhanced CNN to classify Google Street View images into one of three classes: 
\begin{enumerate}
    \item No utility systems
    \item Utility systems without overgrown vegetation
    \item Utility systems with overgrown vegetation
\end{enumerate}
To collect and label our data, we extract images from Google Street View and manually create labels for the ground truth samples. More detail is given in Section~\ref{sec:4}. We take advantage of pretrained models available in Pytorch, using transfer learning by freezing the early layers and resizing and tuning the fully-connected layers on our dataset. The three pretrained pytorch models examined in this work are \textit{AlexNet}, \textit{ResNet18}, and \textit{VGG11}.

While the HOG and Hough transforms, which are described in detail in Section~\ref{sec:4}, are powerful feature extractors, they do not output any information based on pixel color, which is important when identifying power lines against a blue sky, green vegetation, or distinguishing them different color lines on the street. Therefore, this work leverages both the image and the transformed features by stacking them together as the CNN input, making the input size $5\times 224\times 224$ because each transform has one output channel. The first convolutional layer will thus also have to be retrained to learn this size input, but the weights for the first three of the five channels can be initialized from pretrained models. Figure~\ref{fig:net} details this process.

The images collected from Google had a default size of $3\times 640 \times 640$, so they are first resized to $3\times 224\times 224$ to be compatible with pretrained networks. The $1,330$ images are divided into $80\%$ training set, $10\%$ development set, and $10\%$ test set. Then the HOG and Hough transforms are applied to and stacked on each input image to create $5\times 224\times 224$ inputs.

For each network, the loss, $L$, is calculated during training by using the the cross entropy loss function:
\begin{equation}\label{eqn:loss}
    L(y,c) =  w_c \left[ -y_c + \log{\sum_{j=1}^3 \left(\exp{y_j} \right)}  \right] 
\end{equation}
where $y$ is the scores vector of size three (for three classes) generated from the forward pass of the model, $c$ is the correct class identifier (0, 1, or 2), and $w_c$ is the weight for class $c$. The weights $w_c$ can help account for class imbalance in the dataset, and also prioritize accuracy on classes of interest by punishing them more for misclassification. In this study, the weights were set according to the number of samples in each class:
\begin{figure*}[htp]
    \centering
    \includegraphics[width=0.8\linewidth]{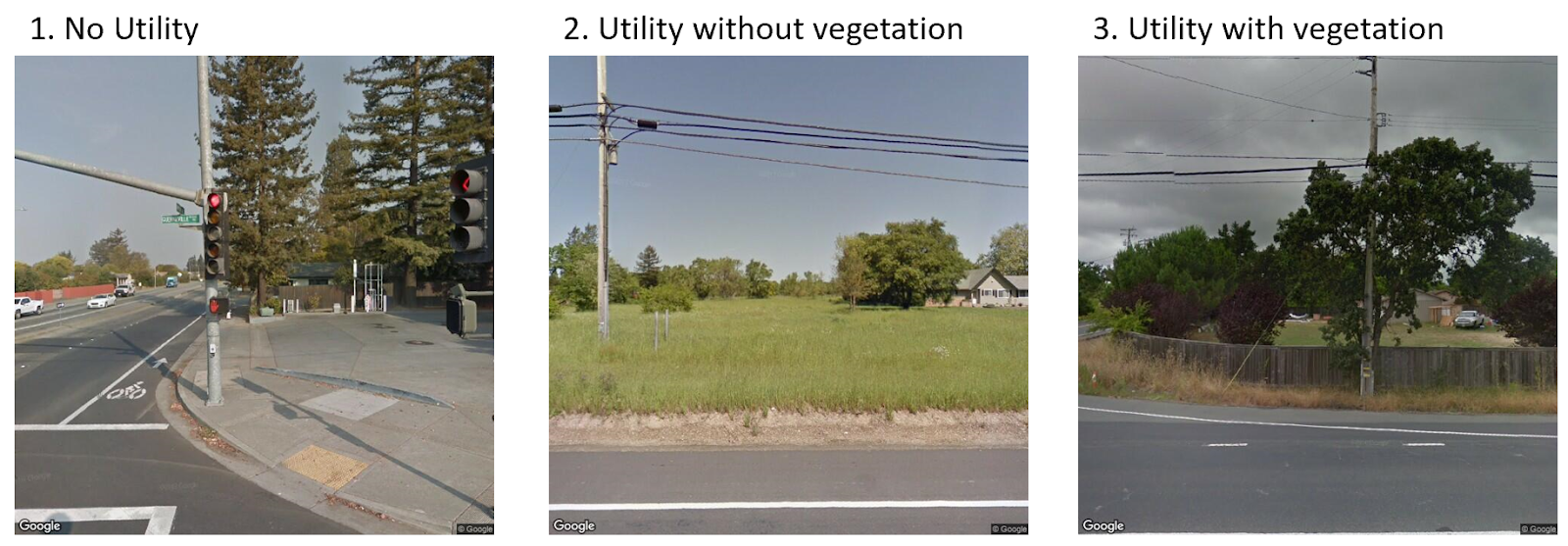}
    \caption{Labeled images from each class.}
    \label{fig:class_examples.png}
\end{figure*}
\begin{equation}
    w_c = N_{c,max}/N_c
\end{equation}
where $N_c$ is the number of samples in class $c$, and $N_{c,max}$ is the number of samples in the largest class. The loss function is minimized using ADAM optimization~\cite{Kingma-2015-Adam}, where the training data is divided into batches, the derivative of the loss function with respect to each trainable weight, $W_k$, and bias, $b_k$, is computed with backpropagation:
\begin{equation}\label{eqn:adam1}
\nabla W_k = \frac{\partial L}{\partial W_{k}}, \quad 
\nabla b_k = \frac{\partial L}{\partial b_{k}}.
\end{equation}
Next, a decaying average of the first and second gradient moments, $m_t$ and $v_t$, is computed for each weight and bias:
\begin{subequations}
	\begin{gather}
	m_{t,W_k} = \beta_1 m_{t-1,W_k} + (1-\beta_1)\nabla W_k, \\
	m_{t,b_k} = \beta_1 m_{t-1,b_k} + (1-\beta_1)\nabla b_k, \\
	v_{t,W_k} = \beta_2 v_{t-1,W_k} + (1-\beta_2)(\nabla W_k)^2, \\
	v_{t,b_k} = \beta_2 v_{t-1,b_k} + (1-\beta_2)(\nabla b_k)^2,
	\end{gather}
\end{subequations}
where $t$ is the batch number, and standard values of $\beta_1=0.9$ and $\beta_2=0.999$ are used. To overcome the fact that $m_t$ and $v_t$ are biased toward zero for early time steps, the moments are corrected by:
\begin{subequations}
	\begin{gather}
	\hat{m}_{t,W_k} = \frac{m_{t,W_k}}{1-\beta_1^t}, \quad \hat{m}_{t,b_k} = \frac{m_{t,b_k}}{1-\beta_1^t}, \\
	\hat{v}_{t,W_k} = \frac{v_{t,W_k}}{1-\beta_2^t}, \quad \hat{v}_{t,b_k} = \frac{v_{t,b_k}}{1-\beta_2^t}.
	\end{gather}
\end{subequations}
Finally, the optimizer updates the weights and biases:
\begin{subequations}\label{eqn:adam2}
	\begin{gather}
	W_{k} = W_k -\alpha_\tau\frac{\hat{m}_{t,W_k}}{\sqrt{\hat{v}_{t,W_k}}+\epsilon}, \\
	b_{k} = b_k -\alpha_\tau\frac{\hat{m}_{t,b_k}}{\sqrt{\hat{v}_{t,b_k}}+\epsilon},
	\end{gather}
\end{subequations}
where $\alpha_\tau$ is a scheduled learning rate for epoch $\tau$ and $\epsilon=10^{-8}$ is used for numerical stability. In this work, a delayed cosine learning rate schedule was used:
\begin{subequations}\label{eqn:lr}
\begin{align}
\alpha_\tau &= \alpha_0, \qquad \tau \leq \tau_{start} \\
\alpha_\tau &= \frac{1}{2} \alpha_0 \left[1 + \cos{\left(\frac{(\tau-\tau_{start})\pi}{\tau_{max}}\right)} \right], \quad \tau > \tau_{start}
\end{align}
\end{subequations}
where $\alpha_0$ is the initial learning rate, $\tau_{max}$ is the number of training epochs, and $\tau_{start}$ is the number of epochs with constant learning rate.

The batch size, initial learning rate, regularization, max epochs, and start epochs are all hyperparameters than are tuned on a separate development set. For each set of hyperparameters the epoch that performed best on the development set, not necessarily the model at the end of training, was evaluated on the test set.

\section{Dataset and Features} \label{sec:4}

\subsection{Data acquisition}

A labeled dataset of utility assets is not readily available online. Instead, the dataset is scraped from Google Street View. Almost every utility asset is built near a road for accessibility, and Google Street View coverage in the United States, especially northern California, is very high. To gather a significant number of images with limited labeling effort, streets that fall mostly into one of the three classification categories listed in Section~\ref{sec:3} are found. Only streets in northern California are chosen in the study, where the use case for this type of utility asset identification and maintenance prioritization is currently strongest. This also ensures relatively similar image features, which is useful for a relatively small dataset.

Each street is manually geotagged with a starting and ending latitude and longitude. A script is then developed to calculate the street trajectory from starting to ending coordinates, then travel along the street and download images at regular intervals. The majority of images are oriented looking forward (heading = 0 , pitch = 0), at maximum resolution ($3\times 640\times 640$), and with a $90^\circ$ field of view. As one can imagine, not all images are perfectly classified based on the broader, ``average'' street-level classification. Not all images on ``utility'' streets had utilities, and even more images on the ``utility'' and ``vegetation overgrowth'' streets were misclassified. Several were completely anomalous, like a Street View image which happened to be inside a restaurant, or a selfie of a family. To improve accuracy, after downloading all images, each are examined and reclassified by hand if necessary.

Examples of images in each of the three classes are shown in Figure 1. In the following subsections, the two transformations used to detect the power lines are described. The resulting images from these two feature descriptors have been stacked with the three RGB channels of the original images and then fed to our training network as the input tensor.

It was very difficult to find accurate images with overgrown vegetation--as it should be if the utility is performing quality  maintenance. Given this limitation, this class of data was augmented by directly searching for pictures of utility equipment involved in fires, utility equipment undergoing vegetation management, and utility equipment that had fallen over and failed. Some of theses images present an extremely challenging classification task for the CNN, because the perspective and zoom level are completely different.

Ultimately, $1,320$ images were scraped and classified. Of these,  $655$ had no utility equipment,  $572$ had utility equipment with no overgrown vegetation, and $93$ had utility equipment with no overgrown vegetation. Data was augmented by horizontally flipping each image to double the dataset size. Vertically flipping images would not be appropriate as a key identifying feature of powerlines is that they are above the ground. Random cropping to augment data for this dataset would also be problematic as the crop may miss the power lines, which are often only a small subset of the image, so no additional augmentation beyond horizontal flipping was performed.

\subsection{Hough Transformation}
\begin{figure}[htp]
\centering

\subfloat{%
  \includegraphics[clip,width=0.54\columnwidth]{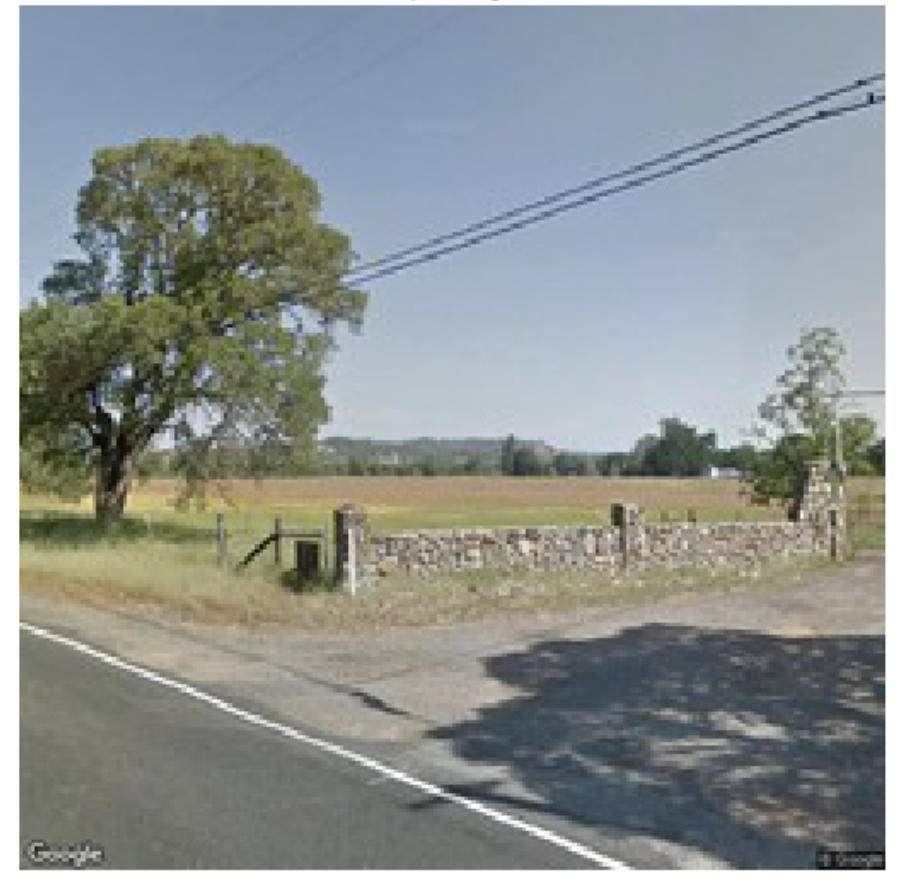}%
}

\subfloat{%
  \includegraphics[clip,width=0.54\columnwidth]{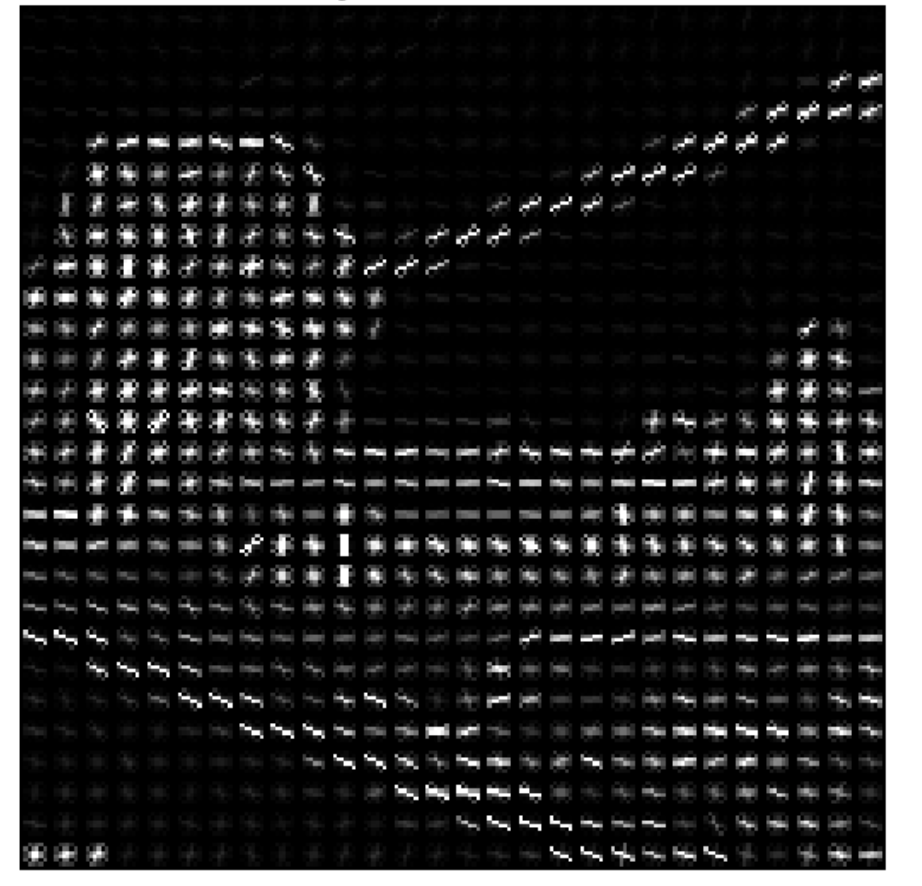}%
  
}

\subfloat{%
  \includegraphics[clip,width=0.51\columnwidth]{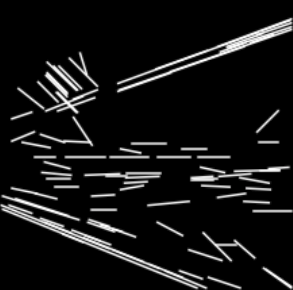}%
}

\caption{(Top to bottom) Input image, HOG and Hough feature descriptors the image}
\label{fig:transforms}
\end{figure}

The Hough transformation (HT), shown in the bottom of Figure~\ref{fig:transforms}, is a method to explore a parameter space and detect straight lines. The classical Hough transform is widely used for the detection of regular curves such as lines, circles, ellipses, etc~\cite{fernandes2008real}. This technique is particularly useful where a global description of a feature(s) is needed and the number of solution classes don't need to be known a priori. The motivating idea behind this transformation technique for line detection is the fact that each input measurement has its contribution to a globally consistent solution (e.g. the physical line which gave rise to that image point). As we know, lines are parameterized as $y= mx + c$ where $m$ is the slope and $c$ is the y-intercept. However, when the line is vertical, the slope goes to infinity. Therefore, a polar coordinate system is preferred for this purpose. To that end, a segment perpendicular to the line and leading to the origin is usually constructed. This line is represented by the length of the segment r, and its angle with the x-axis, $\theta$. Therefore it can be described by the parametric equation,
\begin{align}
    r = x cos \theta + y sin \theta,
\end{align}
where $r$ and $\theta$ are the unknown variables that need to be determined. In other words, the coordinates of points of edge segments $(x_i,y_i)$ in the cartesian image space map to curves in the polar Hough transform space.

The Hough transformation constructs a histogram array representing the parameter space (\textit{i.e.}, a $M\times N$ matrix, for $M$ different values of the radius and $N$ different values of $\theta$). For each parameter combination, $r$ and $\theta$, the number of non-zero pixels in the input image that would fall close to the corresponding line is found. Each non-zero pixel can be interpreted as voting for the potential line candidates. The local maxima in the resulting histogram indicate the parameters of the most probable lines.

Another approach for line detection is the progressive Probabilistic Hough Transform.  The Randomized HT overcomes the shortcomings of HT method such as high computational burden, low detection accuracy, and possibility of missing objects ~\cite{xu1990new}. The Randomized HT works based on random sampling in place of pixel scanning, converging mapping instead of diverging mapping, and dynamic storage in place of accumulation array. Strictly speaking, it is based on the assumption that using a random subset of voting points gives a better approximation, and that lines can be extracted during the voting process by walking along connected components. This returns the beginning and end of each line segment, which is useful.

The probabilistic Hough transform function has three parameters: a general threshold that is applied to the Hough accumulator, a minimum line length and the line gap that influences line merging. 

Before applying the transformation on the street view images (using OpenCV in Python) we employ the Canny Edge Detection to find the edges of the images. The Canny edge detector can produce a set of boundary descriptors for the further image processing. This function requires gray images as input. Therefore, we convert the input images to gray scale images before feeding them to the Canny edge detector.

\subsection{Histogram of Oriented Gradients (HOG)}

In this work, the HOG is used as a feature descriptor along with the Hough transformation. The HOG transform, shown in the middle row of Figure~\ref{fig:transforms}, is widely used to detect objects in computer vision and image processing. This scheme was introduced by Dalal and Triggs~\cite{Dalal-2005-Histograms}. The idea behind the HOG descriptor technique is extracting the most important features and counting occurrences of gradient orientation in portions of an image, which is referred to as the detection window, or region of interest (ROI). Roughly speaking, it is a representation of an image or an image patch that simplifies the image by extracting useful information, \textit{i.e.} the gradient orientations, and throwing away extraneous information. The HOG descriptor computes oriented gradient histograms by global image normalization, calculates the gradient of pixel values in the $x$ and $y$-directions, and finally, computes gradient histograms and normalizes across the ROI.

\begin{figure}[htp]
\centering
  \includegraphics[clip,width=.65\columnwidth]{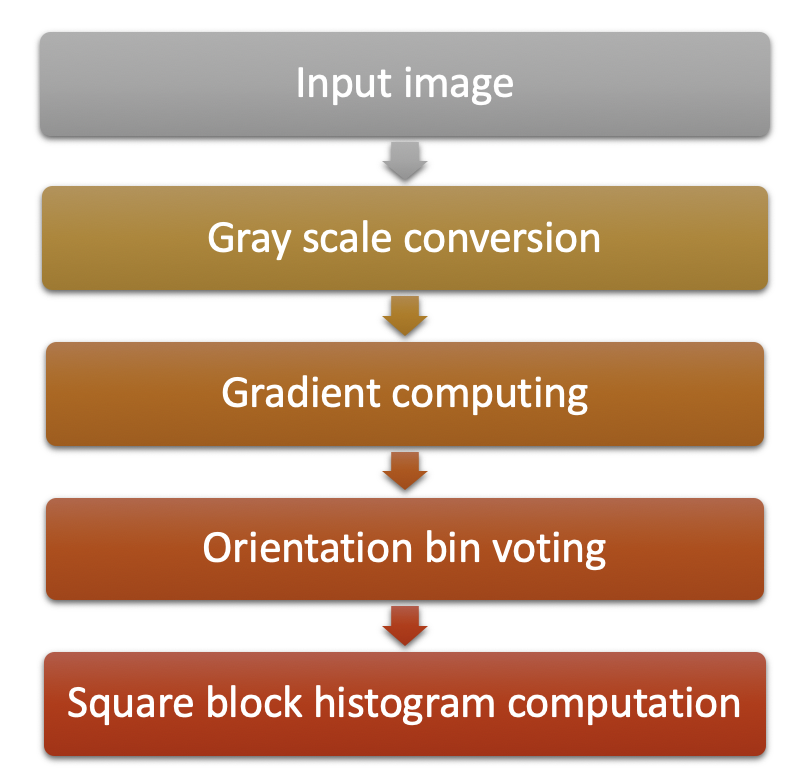}
\caption{The algorithm for Histogram of Oriented Gradients}
\label{fig:hog_alg}
\end{figure}

The following steps describe the HOG descriptor algorithm shown schematically in Figure~\ref{fig:hog_alg}. First, the image is divided into small sub-images called cells or ROIs. for each cell, a histogram of gradient directions or edge orientations is computed. Cells can be rectangular (R-HOG) or circular (C-HOG). Usually blocks overlap each other, so that the same cell may be in several blocks. For each pixel within the cell, vertical and horizontal gradients are obtained. This is done by employing the 1-D Sobel horizontal and vertical operators as following:
\begin{align}
    G_x = Y(y, x+1) - Y(y, x-1)\nonumber\\
    G_x = Y(y+1, x) - Y(y-1, x)
\end{align}
where $Y(y,x)$ is the pixel intensity at coordinates $x$ and $y$. The magnitude and phase of the gradient are determined as:
\begin{align}
    G(y,x) = \sqrt{G_x(y,x)^2 + G_y(y,x)^2}\nonumber\\
    \theta(y,x) = arctan\left(\dfrac{G_y(y,x)}{G_x(y,x)}\right)
\end{align}
The next step is discretizing each cell into angular bins according to the gradient orientation. Certain number of bins are chosen for the angle. Each cell's pixel contributes the weighted gradient to its corresponding angular bin. Subsequently, groups of adjacent cells are considered as spatial regions called blocks. The grouping of cells into a block is the basis for grouping and normalization of histograms. Normalized group of histograms represents the block histogram. The set of these block histograms represents the descriptor. The detector window descriptor is used as information for object recognition. Since different images may have different contrast, contrast normalization would be very useful. Normalization is done on the histogram vector $v$ within a block. Either $L_1$ or $L_2$ can be used for this purpose.

Oriented image gradients are useful because the magnitude of gradients is large around lines, edges, and corners (regions of abrupt intensity changes), which helps improving classification and can be very useful for power line detection. The HOG transform effectively increases the salience of edge orientation while maintaining illumination invariance. This feature simplifies the image by extracting useful gradient distribution information and discarding other texture and color information. Clearly, as seen in Figure~\ref{fig:transforms}, the outputs of both the HOG and HT are not useful for the purpose of human image interpretation, but they are very useful for computer object detection. Support Vector Machines (SVMs) and CNNs trained on HOG outputs in addition to the typical RGB channels for a colored image can outperform vanilla image inputs.

\section{Results and Discussion} \label{sec:5}

The three pretrained models examined in this work are the \textit{AlexNet}, \textit{ResNet18}, and \textit{VGG11}. The loss function used for each model is cross entropy loss given by Equation~\ref{eqn:loss}, and ADAM optimizer described in Equations~\ref{eqn:adam1}~-~\ref{eqn:adam2} is used. The learning rate, regularization, batch size, max epochs and start epochs from Equation~\ref{eqn:lr} are hyperparameters tuned using the development set. The final hyperparameters chosen and the best test set accuracy for each model are shown in Table~\ref{tab:res1}. For all networks, only the fully-connected layers of the final classifier and the very first convolutional layer (to learn weights for a 5 channel input) are trained. All models are trained using the exact same train, development, and test sets.

\begin{table*}[h]
\centering
    \begin{tabular}{|P{1.3cm}|P{1.6cm}|P{1.5cm}|P{1.5cm}|P{1.5cm}|P{2.1cm}|P{1.5cm}|P{1.5cm}|}
        \hline
        \textbf{Model} & \textbf{Batch size} & \textbf{Initial learning rate}, $\alpha_0$ & \textbf{Max epochs}, $\tau_{max}$ & \textbf{Start epochs}, $\tau_{start}$ & \textbf{Regularization} & \textbf{Test set accuracy} & \textbf{Time per epoch} (s)\\
        \hline
        AlexNet  & 64 & 2e-5 & 70  & 30 &  1e-4  &  70.99\%   &  5.1 \\        
        ResNet18 & 96 & 8e-4 & 150 & 50 &  5e-4  &  78.63\%   &  12.0 \\
        VGG11    & 32 & 1e-4 & 11  & 5 &  1e-3  &  80.15\%   &  42.1 \\
        \hline
    \end{tabular}
    \caption{Hyperparameters and best performance for each of the three pretrained models chosen}
    \label{tab:res1}
\end{table*}

Interestingly, development set accuracy is $79.00\%$, $84.33\%$, and $85.07\%$ for AlexNet , ResNet18, and VGG11 respectively, which is quite a bit better than the test set accuracy for each network. This is likely due to the relatively small size of each set even after data augmentation 268 and 262 images in the development and test set. This suggests that more data could further improve accuracy. Nevertheless, the early results are promising.

As shown in Table~\ref{tab:res1}, \textit{AlexNet} is the fastest model to run, but it has the poorest accuracy. On the other hand, \textit{VGG11} achieves the highest accuracy, but runs the slowest. Meanwhile, \textit{ResNet18} exhibits a good balance of both speed and accuracy, so this model is chosen to explore farther.

In the best \textit{ResNet18} model above, the best training and development set accuracy are $97.63\%$ and $84.33\%$ respectively, suggesting that the model is overfitting to the training set. To counteract this, a new model is trained after inserting a dropout layer with $50\%$ dropout probability before the final fully-connected layer. Additionally, to farther improve model flexibility, all layers are unfrozen and allowed to train, but with a learning rate $1,000$ times lower than the learning rate of the first and last layers. This allows for fine tuning of the pretrained intermediate layers to fit the new utility line dataset.

Furthermore, it is noted that while the overall test accuracy was $78.63\%$ for the best \textit{ResNet18} model, the accuracy on the images with encroaching vegetation on the power lines is only $72.22\%$, which is significant because that is the most important class to accurately categorize so that utilities know where they need to perform maintenance. To combat this, the weight for this class, $w_c$ in Equation~\ref{eqn:loss}, is doubled to penalize the network more for misclassifying these images. Note that this is likely to make overall accuracy worse, as more images will be flagged as potentially dangerous. This is acceptable though, because flagged images can always be manually examined to determine if maintenance needs to be provided. This is obviously far better than missing these images and not providing needed maintenance, which could cause an increased fire risk.

\begin{figure}
    \centering
    \includegraphics[width=0.9\columnwidth]{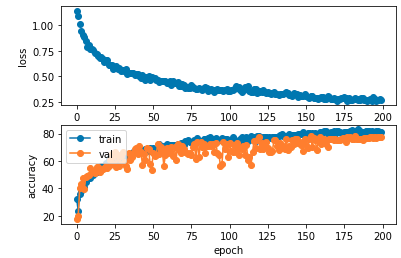}
    \caption{Loss for each epoch during training (top) and classification accuracy for the training and development set (bottom)}
    \label{fig:training}
\end{figure}

Model ensemble is additionally employed at this step to achieve the best possible model. The model is trained for $200$ epochs and saved after each one. The $10$ best models, determined by development set performance, are then applied to the test set. To determine the final classification, each of the $10$ models votes on which class to put each image in, and each image is classified according to the class with the most votes. The loss and accuracy for the training and development sets during this process are shown in Figure~\ref{fig:training}.

Figure~\ref{fig:training} shows that adding the dropout layer successfully overcame the overfitting observed when training the earlier models in Table~\ref{tab:res1}, as the development and training accuracy are very close together throughout the training. The top row shows that the loss behaves as expected for each epoch -- initial loss is roughly $ln(3)=1.09$, it starts to flatten out before decay schedule kicks in at epoch $100$, and then continues to slightly decrease. 

The $10$ best epochs from this model, ensuring that they are at least $5$ epochs apart for substantial differences, are extracted and evaulated on the test set. The maximum test set accuracy for the $10$ models is $73.97\%$. Note that this is worse than achieved in Table~\ref{tab:res1} due to the fact that the weights were changed to predict more vegetation. The best of the $10$ models correctly classified $17$ of $18$ ($94.44\%$) of the vegetation images, while outputting $10$ ``false positives''. Taking the ensemble of the $10$ models and implementing a voting procedure yielded an overall accuracy of $75.95\%$. The confusion matrix for the ensemble model is shown below:
\begin{figure}[!h]
    \centering
    \includegraphics[width=0.9\columnwidth]{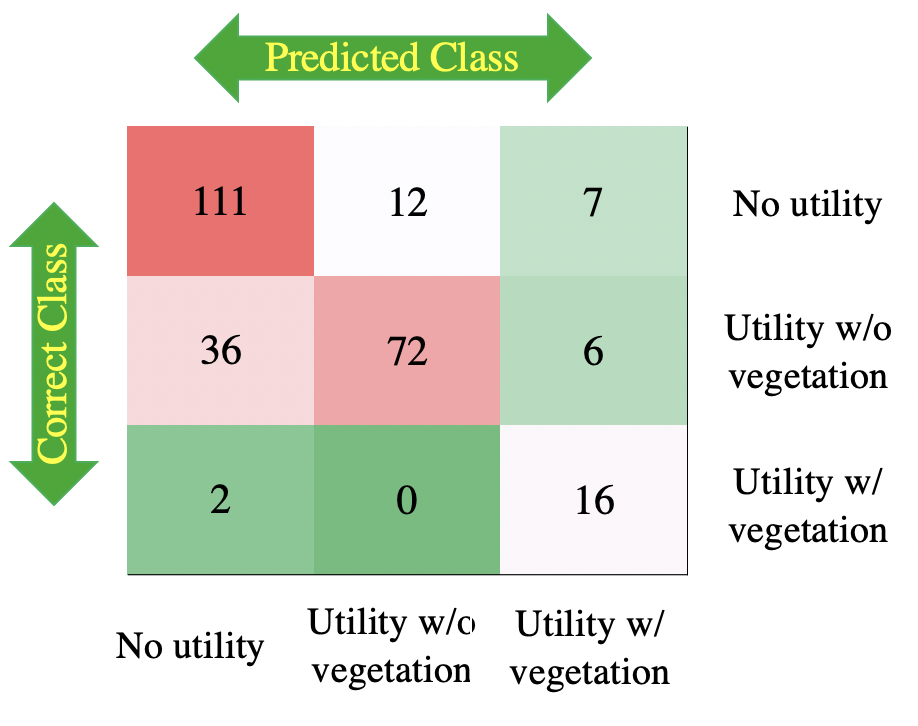}
\end{figure}
This confusion matrix demonstrates that the ensemble model correctly predicts $16$ out of $18$ ($88.88\%)$ of the images labeled as vegetation overgrowth. The two images below show the two that were misclassified, both as ``no utility.''
\begin{figure}[!h]
    \centering
    \includegraphics[width=0.95\columnwidth]{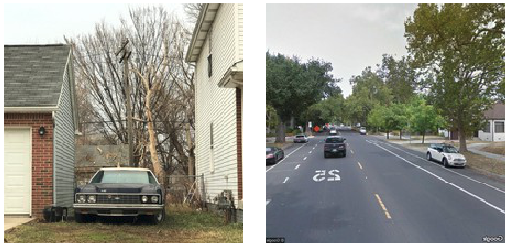}
    \caption{The two images labeled as ``vegetation overgrowth'' that the network misclassified as ``no utility''}
    \label{fig:misclassify}
\end{figure}
\begin{figure}[ht]
    \centering
    \includegraphics[width=0.90\columnwidth]{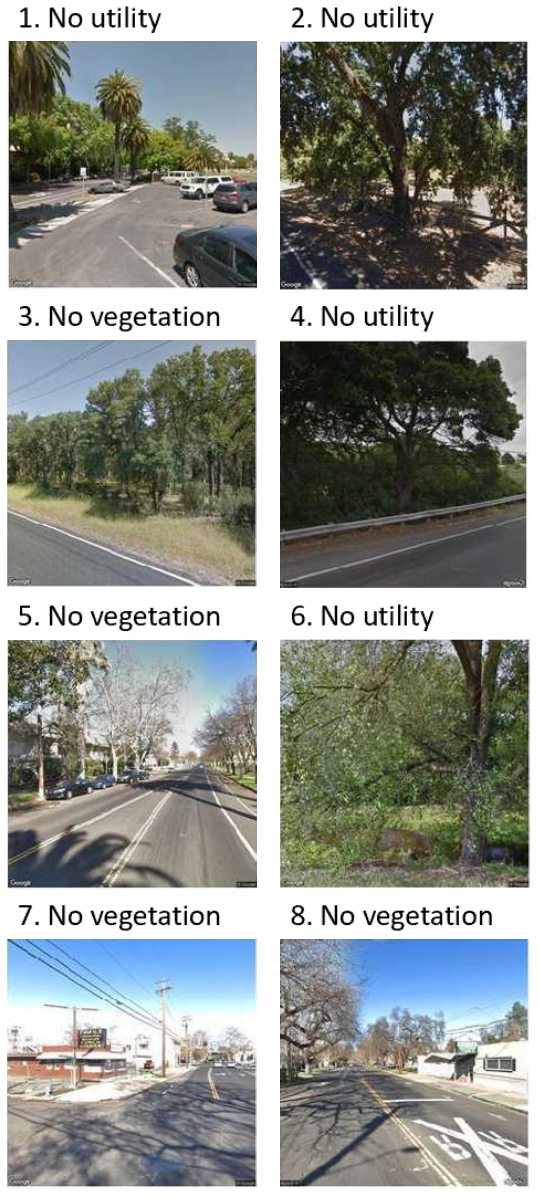}
    \caption{The eight unique false positive images from the ensemble network. The horizontally flipped image of Images 1-5 were also false positives.}
    \label{fig:misclassify2}
\end{figure}

The left image has power lines very well mixed into the tree in the background, but they are very difficult to detect, so it is understandable labeling this image as ``no utility.'' The image on the right actually does not have any powerlines, so this was mislabeled during the manual labeling process, and the network correctly classified it, pointing out our human error. Interestingly, the network classified the horizontally flipped version of each of these images as ``vegetation.''

The confusion matrix also shows that there are $13$ false positives, where the network classified vegetation overgrowth for images labeled without it. This number is quite substantial, but it is expected in the way the weights in the loss function were defined because it is much more dangerous to have false positives than negatives. Because the test set was augmented by horizontally flipping images, only $8$ of these $13$ are unique images as $5$ sets of mirrored images appeared in this group. The images in this group, as seen in Figure~\ref{fig:misclassify2} are also understandable, as many of them have a lot of vegetation, some very close to power lines.

\begin{figure}
    \centering
    \includegraphics[width=0.9\linewidth]{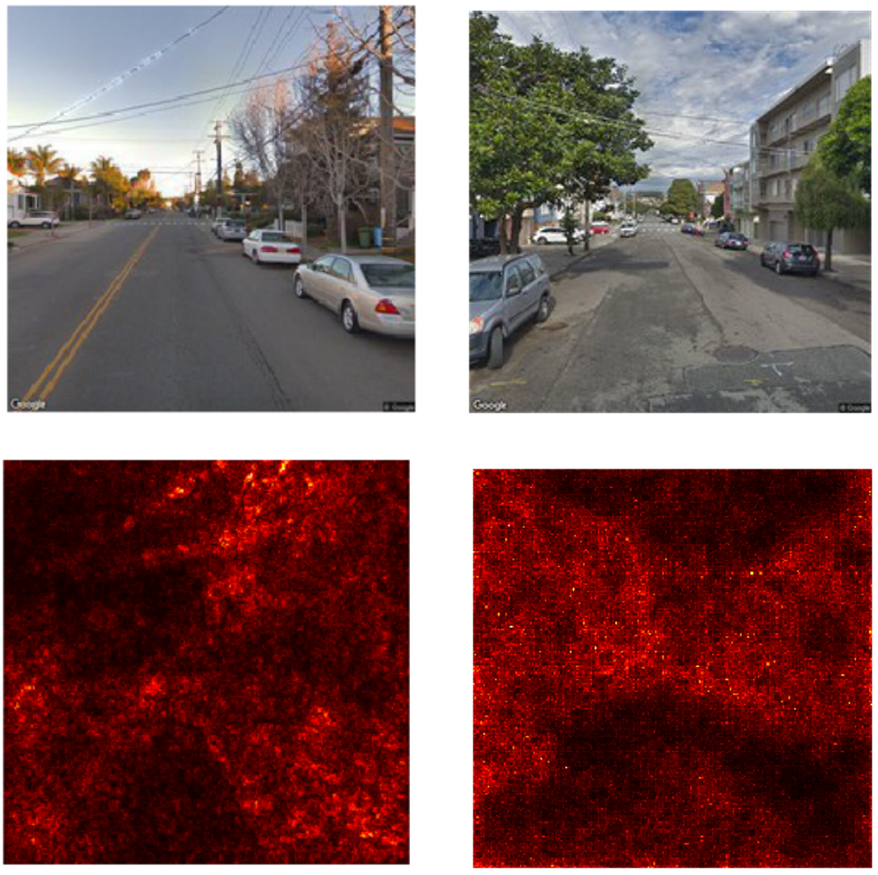}
    \caption{Salience maps for two images in the vegetation class. Brighter red indicates a stronger influence on the classification.}
    \label{fig:sal}
\end{figure}

Salience maps were created to visualize which pixels in various training images were causing major differences in the network's scoring. Brighter red corresponds to greater activation, and all images shown in Figure~\ref{fig:sal} are chosen from the correct category of "Vegetation overgrowth". The images on the left show three separate sets of overlapping power lines all activating \textit{VGG11}, demonstrating that network is in fact learning to recognize power utility assets. It appears the network is also responding to the edge of the street, and the dark-to-light transition in the street. The images on the right display strong activation for the entire area with trees and power lines for \textit{ResNet18}.

\section{Conclusion}
This work developed a framework for identifying utility systems that pose a fire risk using computer vision approaches. Images were scraped from Google Street View. HOG and Hough features were computed and concatenated to the images. A variety of CNN architectures were trained through transfer learning. \textit{VGG11} proved the most accurate on the test set (80.15\%), while \textit{ResNet18} maintained nearly the same level of accuracy with much lower computational cost. 

Future work should include more detailed classes to refine the prioritization, and methods for estimating vegetation distance to utility assets quantitatively. Future work should also focus on obtaining accurate and consistent data--particularly for the class in which vegetation intersects with utility assets.


{\small
\bibliographystyle{ieeetr}
\bibliography{bibliography}
}

\end{document}